# ARE VISUAL DICTIONARIES GENERALIZABLE?

*Otávio A. B. Penatti*[(1)], *Eduardo Valle*[(1,2)], *Ricardo da S. Torres*[(1)*]

(1)RECOD Lab, Institute of Computing, University of Campinas, Brazil

(2)Dept. of Computer Engineering and Industrial Automation, School of Electrical and Computer Engineering, University of Campinas, Brazil

## ABSTRACT

Mid-level features based on visual dictionaries are today a cornerstone of systems for classification and retrieval of images. Those state-of-the-art representations depend crucially on the choice of a codebook (visual dictionary), which is usually derived from the dataset. In general-purpose, dynamic image collections (e.g., the Web), one cannot have the entire collection in order to extract a representative dictionary. However, based on the hypothesis that the dictionary reflects only the diversity of low-level appearances and does not capture semantics, we argue that a dictionary based on a small subset of the data, or even on an entirely different dataset, is able to produce a good representation, provided that the chosen images span a diverse enough portion of the low-level feature space. Our experiments confirm that hypothesis, opening the opportunity to greatly alleviate the burden in generating the codebook, and confirming the feasibility of employing visual dictionaries in large-scale dynamic environments.

## 1. INTRODUCTION

Many of the most effective representations for image classification are based on visual dictionaries, in the so called bags-of-(visual-)words model. That model brings to images many intuitions used previously on textual documents [1], mainly the idea of the use of statistical information on the contents of a document (words or local patches), regardless of its internal structure (phrases or geometry).

Traditionally, a visual dictionary is generated as follows. Local feature vectors are extracted from an image dataset. Popular extraction protocols are based on interest point detectors [2] or dense sampling [3] associated with a description of image patches, often using SIFT-based descriptors [4]. The next step is to cluster the points cloud in the feature space resulting in a feature space quantization, which tends to group points considering their appearance. The clusters in the feature space are the visual words of the dictionary. The images are then represented by a global feature vector which encodes the frequency of occurrence of each visual word. This feature vector is the so-called *bag-of-words* representation.

The whole process of dictionary creation is normally based on images from the same collection that will be represented. In the closed datasets popularly used in the literature [5], like the 15-Scenes, Caltech-101 and 256, and Pascal VOC, the amount of images is fixed, therefore no new content is added after the dictionary is created. However, in a large-scale dynamic scenario, like the Web, images are constantly inserted and deleted. In order to represent well those collections, how should a dictionary be created? Sampling the entire collection and updating it often is unfeasible. Therefore, we would like to create the dictionary from a small fraction of the images and update it seldom (or never). How much would that degrade the accuracy of the representation?

We argue that the impact will be lesser than it might be expected. The term “dictionary” is somewhat a misnomer, because it is not concerned with semantic information. More often than not, the creation of the visual dictionary ignores completely the image labels, which capture the users’ conceptual view of the images, and uses only the low-level features. Provided that the selected sample represents well enough that low-level feature space (being, for such, diverse in terms of appearances), the dictionary obtained will be sufficiently accurate, even if based on a small subset of the collection, or even on a completely different collection.

We have used the Caltech-101 and the 15-scenes datasets in order to evaluate the impact of using “cross-base” dictionaries, i.e., dictionaries created from samples of one dataset to create the bags-of-words of the images in another dataset. We have also used the Caltech-101 dataset alone to evaluate the impact of diversity on the quality of the dictionary used.

## 2. VISUAL DICTIONARIES

The bag-of-words model starts with the generation of a visual dictionary, which is used as a codebook to encode the low-level features. The objective is to preserve the discriminating power of local descriptions while pooling those features into a single feature vector [6].

In order to generate the dictionary, one usually samples or clusters the low-level features of the images. Those can be based on interest point detectors, or employ dense sampling in

*Authors thank to Fapesp (Grant Nos. 2009/10554-8, 2009/18438-7), Capes, and CNPq.

a regular grid, the latter method being preferred for classification [3]. Each of the points sampled from an image generates a local feature vector, SIFT [4] being an usual choice. Those feature vectors are then either clustered or randomly sampled in order to obtain a number of vectors (ranging from a few hundred to several thousands, depending on the application) to compose the visual dictionary [7, 8]. Although clustering (often k-means) is still a popular choice, due to the curse of dimensionality, the simple random sampling of points has been shown to generate dictionaries of similar quality, at a much lesser cost [7].

Image representations are generated based on the created dictionary. The first step is to assign one or more of the visual words to each of the points in an image. Popular approaches are based on *hard* or *soft* assignment, with an advantage to the last one [9–11]. Soft assignment of a point $p_i$ to a dictionary of $k$ words can be formally given by Equation 1 [10]:

$$\alpha_{i,j} = \frac{K_\sigma(D(p_i, w_j))}{\sum_{l=1}^{k} K_\sigma(D(p_i, w_l))} \quad (1)$$

where $j$ varies from 1 to $k$ and $K_\sigma(x) = \frac{1}{\sqrt{2\pi} \times \sigma} \times exp(-\frac{1}{2}\frac{x^2}{\sigma^2})$. The next step is the pooling over the image points. The traditional histogram of visual words can also be called the *average* pooling over the points. Another pooling method with better results is the *max* pooling [6]. The pooling step generates the final image feature vector that can be used in the desired application. Max pooling can be formally defined by Equation 2:

$$h_j = \max_{i \in N} \alpha_{i,j} \quad (2)$$

where N is the number of points in the image and j varies from 1 to $k$.

In the literature, we can find many works evaluating steps or proposing improvements on the dictionary creation or use. Some papers evaluate the use of different detectors and descriptors [3]. Others evaluate the effect of assignment methods [9, 10, 12] and also the different pooling strategies [6]. There are also many attempts to encode spatial information on the bag-of-words representation [13, 14].

Another paper shows the dataset bias [5] by a classification setup, training on images from one dataset and testing on images of another dataset. Those experiments are somewhat similar to our experiments in this paper, however they work on the classification level, while we focus on the representation level. Furthermore, many works do not consider the possible use of visual dictionaries in dynamic environments, such as the Web. Our focus in this paper is to evaluate how the source of information impacts the quality of dictionaries.

## 3. EXPERIMENTAL SETUP

As pointed in the introduction, the traditional datasets used in the experiments of literature are static, which means that no new images appear or disappear after the dictionary creation. However, in a Web scenario, where new content is constantly being indexed (while others are being deleted), is the previously created visual dictionary still good for representing the new images? Of course, regenerating the dictionary whenever the database changes is unfeasible. Therefore, we try to explore those aspects going in the direction to evaluate if it is feasible to use visual dictionaries in this dynamic environment. The paper addresses the following questions:

- Are dictionaries created over certain images generalizable to images of other nature? (if we create the dictionary over images of certain nature, will it represent well images of other nature?)

- Do we need to have a representative subset of the whole collection to create a good dictionary?

To answer each of those questions we performed several experiments. In all experiments the parameters for dictionary generation and image representation are the same: dense sampling (6 pixels) [3] and SIFT descriptor, 1000 visual words selected by random, and soft assignment ($\sigma = 60$) with max pooling (Equations 1 and 2 from Section 2). Those are one of the best parameters configurations found in literature [6]. We have used two popular datasets, 15-Scenes and Caltech-101.

We have conducted experiments in a classification protocol, with SVM using linear kernel. A balanced validation was performed, varying the number of training samples per class and using the rest of images in the test set. The training samples were randomly selected and each training and classification was performed $5$ times. We have evaluated the results in terms of classification accuracy.

## 4. RESULTS AND DISCUSSION

The experimental results and discussion are shown considering each of the two questions just presented.

### 4.1. Are dictionaries generalizable?

To answer the first question, we have created $5$ dictionaries based on each of the two datasets. Then, we have used each dictionary to represent images from the same dataset and images from the other dataset. For example, in one of the cases, we have represented images of 15-Scenes using a dictionary created over Caltech-101 images. The hypothesis is that images whose representations are based on a dictionary created over images of the same dataset are better than representations based on dictionaries created over images of the other dataset.

Figure 1(a) compares the classification accuracies when the images from 15-Scenes dataset are represented by dictionaries either based on their own images or based on Caltech-101 images. We can see that there is no statistical difference for any of the training set sizes considered. This phenomenon goes against our hypothesis, showing that, to represent 15-Scenes images, the dictionary generated over Caltech-101 images is as good as the dictionary based on 15-Scenes images.

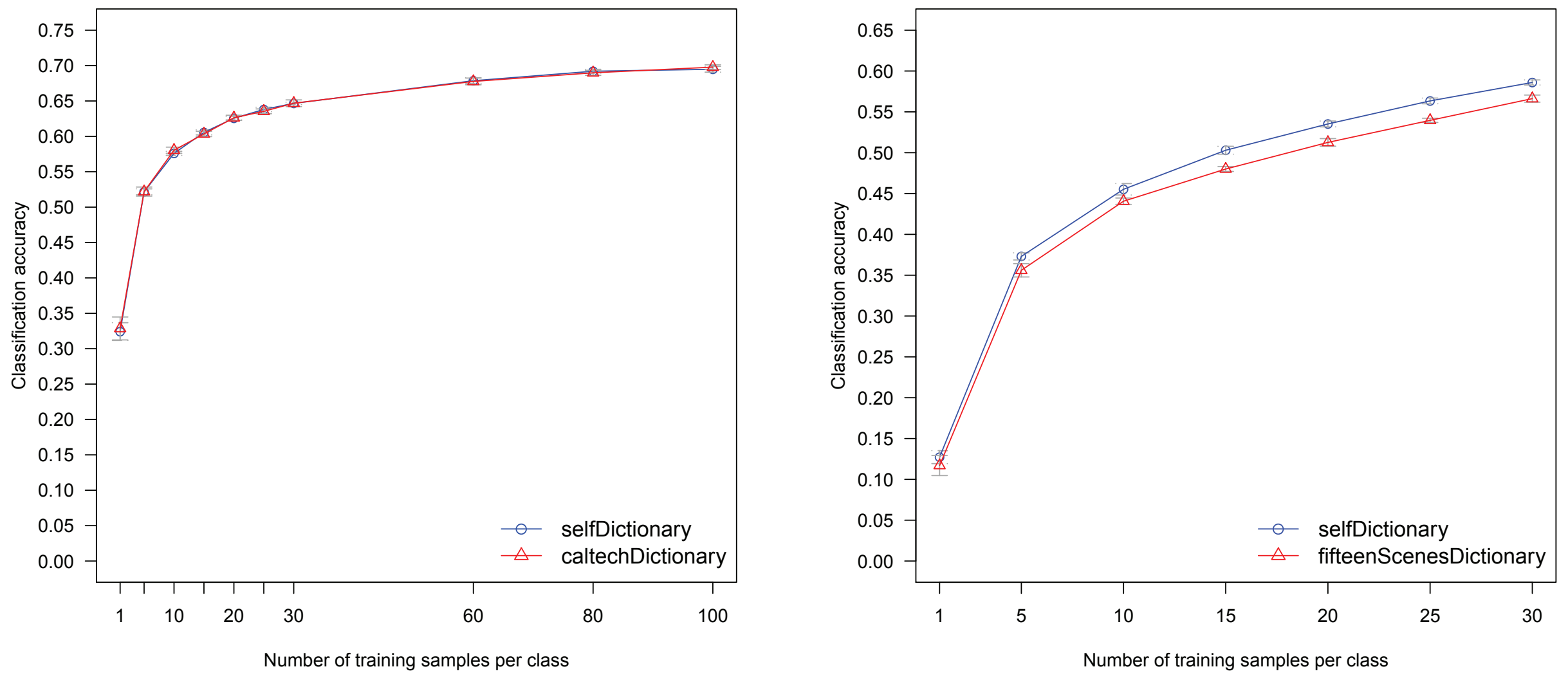

(a) 15-Scenes (b) Caltech-101

**Fig. 1**. Classification accuracies on the datasets using dictionaries based on the same dataset (blue circles) and the other dataset (red triangles). The confidence intervals (error bars) are for $\alpha = 0.05$, on an average of $5$ runs obtained on different dictionaries. In (a) it is seen that the 15-Scenes dataset with its own dictionariy is not significantly better than that using Caltech-101 dictionary. The opposite configuration (b), using 15-Scenes dictionary on Caltech-101 dataset shows some loss of accuracy. Contrarily to Caltech-101, the visual diversity of 15-Scenes is more limited.

Figure 1(b) compares the classification accuracies when the images from Caltech-101 dataset are represented by dictionaries based on their own images and by dictionaries based on 15-Scenes images. Contrarily to the just presented results for the 15-Scenes dataset, the representations of Caltech-101 images are better if they use dictionaries based on their own images. Although it exists, the difference in favor of Caltech-101 dictionaries is very small.

We can conclude that 15-Scenes images are less variable than Caltech-101 images in terms of SIFT descriptions. The SIFT descriptions of Caltech-101 seem to comprise more of the whole SIFT space, while the SIFT descriptions of 15-Scenes may concentrate only on portions of that space. Another possibility is that the space comprised by Caltech-101 descriptions is larger and covers the space of 15-Scenes descriptions. Therefore, the dictionary based on Caltech-101 is more general than the dictionary based on 15-Scenes images.

Those results answer our first question. The variability of the SIFT descriptions of a dataset is important to indicate how general is a dictionary created over its images. An stereotyped dataset will generated good dictionaries only for itself or for other datasets with the same characteristics. A heterogeneous dataset in terms of feature descriptions can generate dictionaries more widely useful.

It is important to highlight that we are not analyzing if any of the datasets is biased in terms of classes or images. We are providing results indicating the dataset variability in terms of feature space.

With the results presented, we can say that if we use a good dataset in terms of feature space variability, we can generate a dictionary able to represent well many different types of images, even images that are not known yet, like in a Web scenario. Therefore, this is an indication of the possibility to use visual dictionaries in heterogeneous and dynamic environments.

### 4.2. Do we need to have a representative subset of the whole collection to create a good dictionary?

To answer our second question, that raises the necessity of having or not a substantial part of the dataset to generate a good dictionary for representing images, we have prepared an experimental setup varying the number of image classes used for dictionary generation. We have used Caltech-101 as the features source, due to its variability presented in the experiments in the previous section.

We have created $6$ different dictionaries using, in each of them, a different number of classes from Caltech-101. The first dictionary was generated based on images of only $1$ class. The second dictionary was based on images of $6$ classes, which includes the class used in the first dictionary. The following dictionaries kept the incremental aspect, increasing number of classes to $12$, $25$, $50$, and $101$. For each of the dictionaries, we have represented the whole Caltech-101 dataset and have conducted classification experiments in the same fashion presented before.

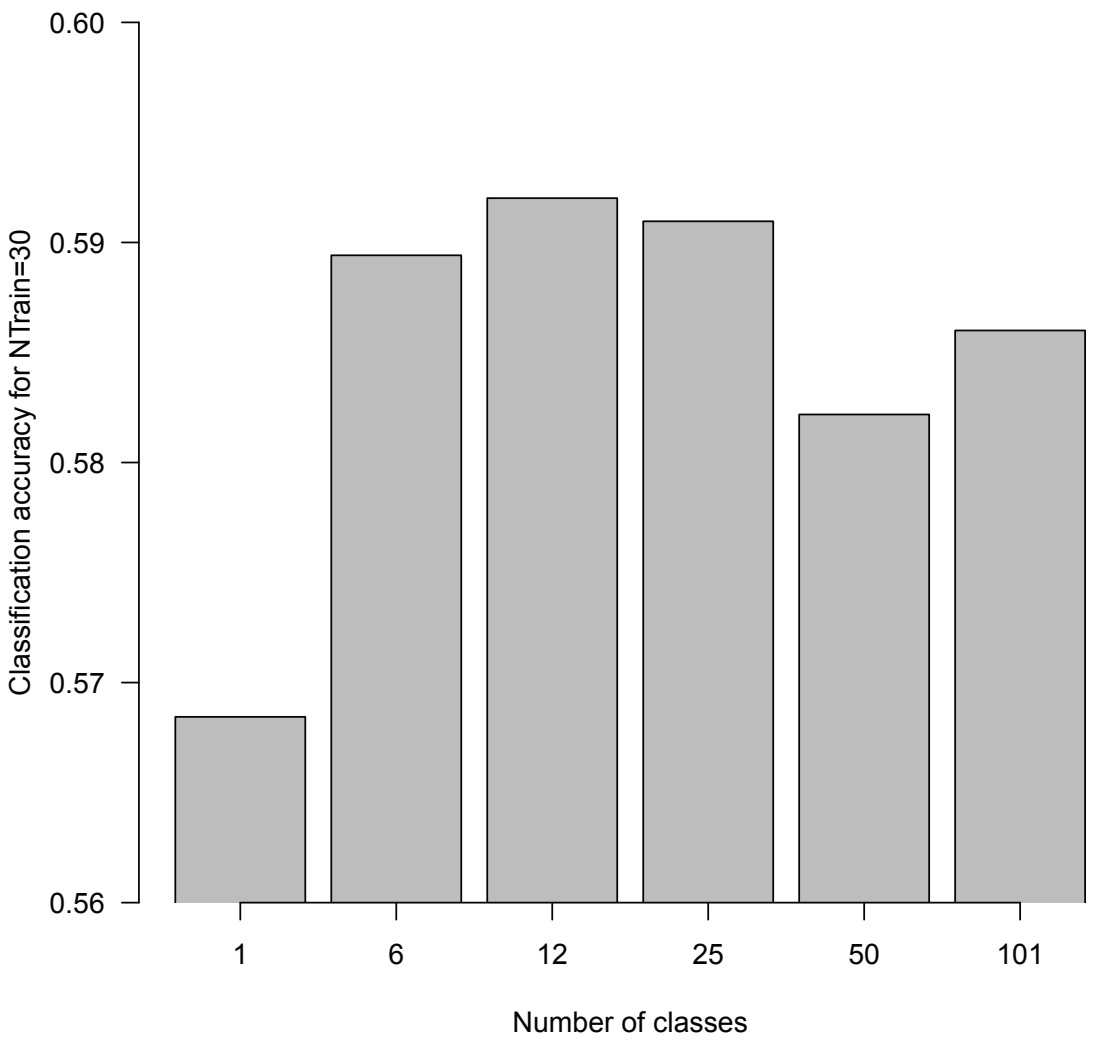

**Fig. 2**. Classification accuracy on Caltech-101 over dictionaries sampled from a varying the number of its classes. Although the results show some random fluctuation, it is clear that as soon as the *visual* diversity becomes acceptable, the accuracy reaches its asymptotic value, even if *semantically* (in terms of label diversity), the sample is still very poor.

Figure 2 shows the accuracies for the 6 different dictionaries when 30 training samples per class are used (NTrain=30) for Caltech-101 dataset.

We can see that the largest difference occurs for the dictionary based on 1 class, and this difference is still around only 2%. Using just 6 classes is enough to generate a good dictionary. Therefore, we can also answer our second question. The results just presented are a good indication that, even with a small portion of the dataset, we can generate a good dictionary. As the low-level descriptor (SIFT) is based on image local textures and not on semantics, the fast dictionary generalization occurs if we have a set of images rich enough in terms of textures, which will cover all the feature space without requiring the use of all images classes. In a Web environment, it should be feasible to use just a small set of images to create a good dictionary that can be later used for representing a large amount of other images.

## 5. DISCUSSION

This paper has evaluated the feasibility of using visual dictionaries in scenarios where the entire dataset is not available for the dictionary construction as, for example, in large-scale dynamic datasets, like the Web. The experiments conducted confirm our hypothesis that dictionaries based on a subset of the collection, or even on a entirely different collection, may still provide good performance, on the condition that the selected sample is visually diverse. A heterogeneous collection in terms of low-level descriptors is able to generate widely useful dictionaries.

Those findings open the opportunity to greatly alleviate the burden in generating the codebook, since, at least for general-purpose datasets, we show that the dictionaries do not have to take into account the entire collection, and may even be based in another small collection of well chosen, visually diverse images.

We would like to explore if those findings also work on special-purpose datasets, in applications like quality control in industry images or aided diagnostics in medical images. We suspect that for those datasets, the selection of a specific dictionary might be more critical than for general-purpose datasets.